\documentclass[10pt, conference, compsocconf]{IEEEtran}

\usepackage{graphicx}
\usepackage{amsmath}
\usepackage{pifont}
\usepackage{float}
\usepackage{hyperref}
\newcommand{\cmark}{\ding{51}}%

\ifCLASSINFOpdf
\else
\fi

\begin{document}
%

\title{Diffusion Dataset Generation: Towards Closing the \textit{Sim2Real Gap} \\ for Pedestrian Detection}



\author{\IEEEauthorblockN{Andrew Farley}
\IEEEauthorblockA{
16amf8@queensu.ca}
\and
\IEEEauthorblockN{Mohsen Zand}
\IEEEauthorblockA{
m.zand@queensu.ca }
\\
\IEEEauthorblockN{
RCVLab, Ingenuity Labs,
Queen's University,
Kingston, Canada
}
\and
\IEEEauthorblockN{Michael Greenspan}
\IEEEauthorblockA{
michael.greenspan@queensu.ca}
}

\maketitle

\begin{abstract}
We propose a method that augments a simulated
dataset using diffusion models
to improve the performance of 
pedestrian detection in real-world data.
The high cost of collecting and annotating data in the real-world
has motivated the use of simulation platforms to create training datasets. 
While simulated data is inexpensive to collect and annotate, it unfortunately does not always closely match 
the distribution of real-world data, 
which is
known as the \textit{sim2real gap}. 
In this paper we propose a novel method of synthetic data creation meant to close the \textit{sim2real gap} 
for the challenging pedestrian detection task. Our method uses a diffusion-based architecture to learn a real-world distribution which, once trained, is used to generate datasets. We mix this generated data with simulated data as a form of augmentation and show that
training on
a combination of generated and simulated
data
increases average precision by as much as 27.3\% for pedestrian detection models in real-world data,
compared against training on purely simulated data.

\end{abstract}


\IEEEpeerreviewmaketitle




%

\section{Introduction}

A well known challenge of supervised Deep Learning methods is that they require large
amounts of labelled data,
which can be costly to acquire.
In the field of pedestrian detection, for example, datasets consisting of street footage coupled with labelled bounding boxes of pedestrians and other scene elements have been constructed~\cite{citypersons},
in which 
$\sim\!\!35$k bounding boxes were manually 
localized and annotated
for 
$\sim\!\!5$k frames of video footage.
Recent advances in human activity recognition ~\cite{human_prediction} and trajectory prediction ~\cite{trajectory_pred} have created a need for more specific datasets that include pedestrians in particular poses and executing distinct activities, which can be relatively rare to identify in real datasets.
An example could be a pedestrian talking 
distractedly
on their phone while crossing a street,
which while a rare occurrence in a natural datasets, may nevertheless be a valuable event to detect.

\begin{figure}
  \centering
  \includegraphics[width=0.5\textwidth]{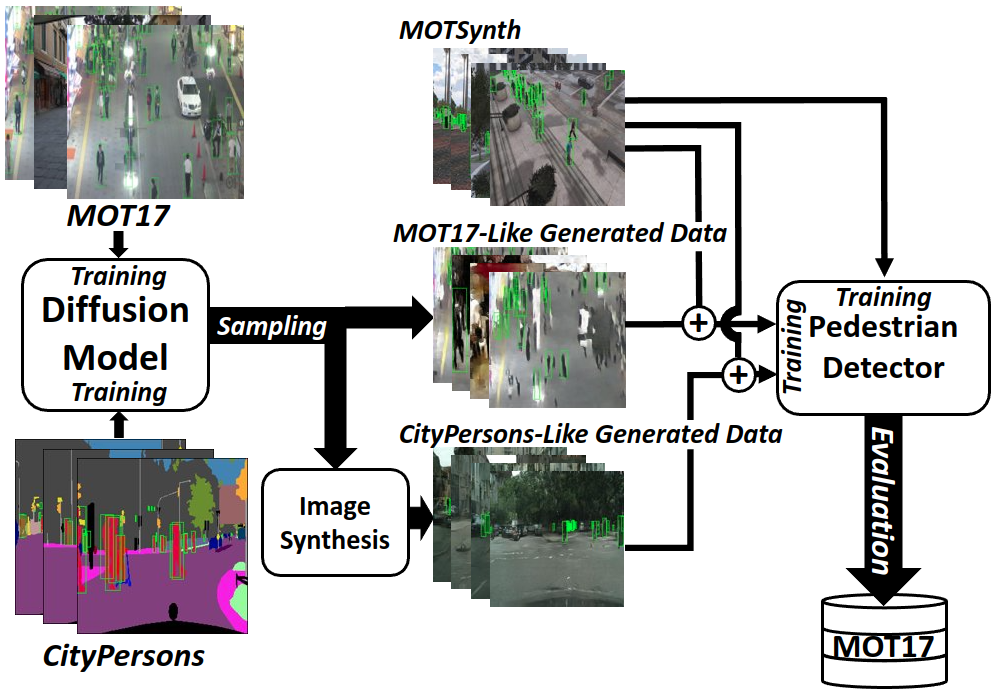}
  \caption{Diffusion data generation for augmentation of pedestrian detection dataset}
  \label{fig:overall_pipeline}
\end{figure}

Recently, simulation has been proposed to address the limitations of real data, and the downsides of collecting and annotating data manually. Video games such as Grand Theft Auto V~\cite{motsynth} and purpose designed simulation platforms such as CARLA~\cite{CARLA}, have been used to generate vehicle-centric, automatically labelled street scene datasets,
which significantly reduces the time and cost associated with dataset creation,
while also potentially increasing the variety and frequency of conditions and events that can occur in a dataset.

One possibility is to use purely simulated data to train a network to then execute on real data. It has been shown that in some cases, networks trained on simulated data can outperform those trained on real data, albeit when tested on different real distributions than those trained on~\cite{motsynth}. A more common scenario is that training on simulation results in a performance drop, due to differences in the simulated and real distributions, which is known as the \textit{sim2real gap}~\cite{pedaugcp, attentionpedaug}. One option is to try and reduce this difference, by making the simulation more closely match (i.e. appear more like) the real images. This is itself a difficult task, as it is not always obvious which aspects of the appearances are most important to a downstream detection task.

\begin{figure}
  \centering
  \includegraphics[width=0.4\textwidth]{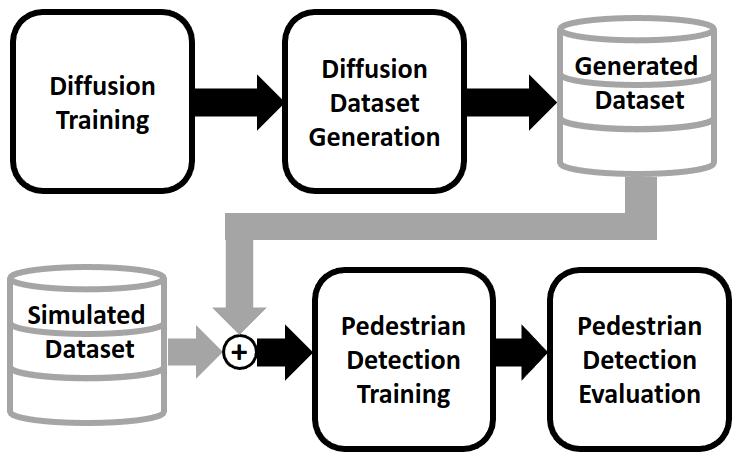}
  \caption{Proposed pipeline for diffusion dataset generation and augmentation}
  \label{fig:general_pipeline}
\end{figure}

An alternative which we propose here is to rely upon a generative process to match the real data distribution and narrow the sim2real gap. Recently, diffusion models have been actively investigated in the generative model literature due to its demonstrated success
and promise in generating data when compared to GANs~\cite{GAN} or VAEs~\cite{VAE}. Diffusion models work by gradually adding noise to an image
to the point that it degrades into pure Gaussian noise,
and then 
iteratively
denoising
the Gaussian.
Once trained, diffusion models can be sampled by passing random Gaussian noise to the learned parameters, resulting in an image 
that while distinct from the initial training set,
is nevertheless representative of the domain that it was trained on. 
To leverage diffusion models, we introduce a novel architecture which is able to learn to produce annotations as well as images, for use in real-world pedestrian detection datasets. This generated data is then used to augment a simulated dataset, resulting in better performance on real-world data. The overall pipeline can be viewed in Figure \ref{fig:overall_pipeline} with a general representation seen in Figure \ref{fig:general_pipeline}.




As the world of machine learning continues to evolve, more and more data will be needed,
incurring larger costs,
and simulated data will continue to 
play a role
to supplement real dataset creation.
Based on the current literature, however,
there appears to be an upper limit on how effective this data might be used for real-world applications. We believe that generative processes, such as our proposed method, will be crucial to closing the \textit{sim2real gap}. As work like this progresses, older real-world datasets will be able to be leveraged to generate new realistic datasets, 
with an insignificant cost 
compared against traditional data collection methods. This generated data can be used in tandem with simulated data for better performance than training on purely simulated data.

The contributions of this paper are as follows:
\begin{itemize}
    \item A novel pipeline of diffusion data generation which simultaneously generates
    both images and bounding box annotations, to mimic real-world pedestrian detection datasets;
    \item A novel method of augmenting simulated datasets with generated data for improved pedestrian detection;
    \item We release our generated datasets into the public domain, to facilitate its use in future research. These datasets are available here: \url{https://github.com/And1210/Diffusion-Generated-Datasets}.
\end{itemize}


\section{Related Work}
\subsection{Simulated Datasets}
A recent work, published in 2021, describes a synthetic dataset collected in the Grand Theft Auto V (GTA V) video game~\cite{motsynth}. This dataset is known as MOTSynth, and is the baseline we will try and improve upon in this paper. 
In their work, Fabbri et al.~\cite{motsynth} aim to answer the question \textit{``Can we advance state-of-the-art methods in pedestrian detection and tracking using only synthetic data?''}. They explored this by using GTA V to simulate different street scenes and implemented a stationary camera to collect a series of video clips with corresponding bounding box frame annotations~\cite{motsynth}. They then trained pedestrian detectors on different subsets of the collected data and tested these models on the real datasets MOT17 and MOT20, 
both of which are part of the annual MOTChallenge~\cite{motsynth}. The collected data is made to match
the distribution of MOT17 and MOT20 by simulating crowded pedestrian scenes as seen in these real-world datasets.

Using the MOTSynth dataset, Fabbri et al.~\cite{motsynth} compared collected results to a pedestrian detection model trained on the COCO dataset and tested on real MOT17 and MOT20 data, and showed an increase in performance by using purely simulated data. In this way, they compared the performance of synthetic data vs. a general purpose real-world dataset. Additionally, Fabbri et al. performed this experiment on person re-identification and pedestrian tracking. Their work showed promising results, reporting a best Average Precision of 78.98\% and 53.90\% when trained on MOTSynth and tested on MOT17 and MOT20 respectively, while their comparative results on COCO training are 76.68\% and 43.67\% respectively~\cite{motsynth}.

Two other simulated datasets are VIPER~\cite{VIPER} and JTA~\cite{JTA}. VIPER was created in 2017 and was made as a benchmark suite for visual perception tasks. Being one of the first, realistic, simulated datasets, they were able to collect a large quantity of data with multiple annotations including semantic segmentation maps, dense pixel-wise semantic instance segmentations, and instance-level semantic boundaries.
VIPER was also collected in GTA V and Richter et al.~\cite{VIPER} provided valuable insights into how they manipulated the game to gather the different kinds of information provided~\cite{VIPER}.

JTA was created in 2018 to address the lack of data in multi-people object tracking with varying occlusion in urban scenarios. Also collected in GTA V, JTA contains much more data when compared to VIPER. JTA collected annotations for 3D data, occlusion labels, tracking information, and pose estimation. The data was collected from a driver, cyclist, and pedestrian point of view and was gathered in a variety of daylight and weather conditions.

\begin{table} 
\centering
\caption{The size and annotations of three synthetic datasets}
\begin{tabular}{c|c|c|c}
    \hline
    \textbf{Dataset} & \textbf{VIPER~\cite{VIPER}} & \textbf{JTA~\cite{JTA}} & \textbf{MOTSynth~\cite{motsynth}} \\
    \hline
    \# Frames & 254k & 460k & 1,382k \\
    \# Instances & 2,750k & 15,341k & 40,780k \\
    3D & \cmark & \cmark & \cmark \\
    Pose & - & \cmark & \cmark \\
    Segmentation & \cmark & - & \cmark \\
    Depth & - & - & \cmark \\
    \hline
\end{tabular}
\label{table:dataset-stats}
\end{table}

The size, and different annotations, of data for MOTSynth, VIPER, and JTA can be viewed in Table \ref{table:dataset-stats}. MOTSynth is by far the largest dataset and also contains the greatest variety of data
annotations. 


\subsection{Synthetic Dataset Methods for Pedestrian Detection}
There have been a number of ways in which synthetic data has been used in the field of pedestrian detection which often involve the use of generative models to close the \textit{sim2real gap}. Kieu et al~\cite{vis2therm}, for example, employ GANs to generate data for the task of pedestrian detection in the thermal domain. Their ``vis2therm'' GAN, based on LSGAN~\cite{lsgan}, learns from the training set to produce extra thermal data which is then used to augment their original training set~\cite{vis2therm}. By using a dataset consisting of 90\% real data and 10\% generated data they show a 9.6\% miss rate reduction when compared to the state-of-the-art~\cite{vis2therm}.

Liu et al.~\cite{APGAN} take a step away from simulation and look to use pedestrians from alternate real-world datasets to augment their primary one. Their method, named APGAN~\cite{APGAN}, first embeds pedestrian crops into a target scene, and then performs a style transfer to ensure the pedestrians appear as though they fit into the target dataset. Using this method they are able to improve the generalization capacity of a pedestrian detection model to other datasets. 

In addition, simulated data has the advantage of creating scenes for specific and somewhat rare situations (e.g. children playing in the street) which are often missed in real data. Huang and Ramanan~\cite{adversarial_imposters} tackle this concept by using a game engine to generate pedestrians in these situations and then using a GAN-based architecture to both discriminate and select realistic images out of their simulated data, named Adversarial Imposters, and to generate more realistic looking pedestrians. They name the result of their work the Precarious Pedestrian Dataset and show that using this dataset with their collected real world examples leads to a 24\% reduction in miss rate compared to their baselines~\cite{adversarial_imposters}.

Nie et al.~\cite{synposes} tackled this issue slightly differently by generating humans in different poses which then were embedded in the scene. They utilized the SMPL model~\cite{SMPL}, coupled with a random latent sampler, to generate different poses. Then, they used rendering software to make these people look realistic. Finally, they embedded the generated people into a scene in a real dataset. In this way they were able to create much more data for corner cases of pedestrian poses. Using this method, Nie et al. showed a 7.7\% AP improvement when compared to their original baseline.

\subsection{Diffusion-Based Generation Methods}
Recently, diffusion has been explored as a new kind of generative model and has been shown to outperform GANs~\cite{GAN} and VAEs~\cite{VAE}. Researchers have begun to explore diffusion as a data augmentation method. Trabucco et al.~\cite{semantic_diffusion_aug} uses an off-the-shelf diffusion model to manipulate the semantics of images. In this way, they were able to take an image of a truck, for example, and randomly change its paint-job, window formation, and chassis structure along with many other attributes providing a much more diverse image augmentation technique. 

Pinaya et al.~\cite{diffusion_brain} explored dataset generation using diffusion models on high-resolution 3D brain scans. They conditioned their model on person attributes such as age, sex, and brain structure volumes which allowed control over the generated images on sampling. In this work, they showed that diffusion created much more realistic images when compared to GANs and provided a synthetic dataset of 100,000 generated images for the community to use.

Akrout et al.~\cite{diffusion_skin} looked to diffusion to generate images for skin condition classification. To achieve this they trained a diffusion model, coupled with CLIP latents gathered from text prompts, on an existing skin condition dataset which allowed them to generate more images using their own text prompts. They showed that using purely synthetic data during training resulted in similar classification accuracy when compared to the real dataset and that mixing these datasets generally provided the best results.

\section{Background}
 The Denoising Diffusion Probabilistic Model (DDPM) is a recent advancement in the field of image generation. On each training iteration there is a forward and backward diffusion process. On forward diffusion, an input image is noised 
 as follows:

\begin{equation}
    \mathcal{N}(x; \mu, \sigma) = \frac{1}{\sigma\sqrt{2\pi}}\exp(-\frac{1}{2}(\frac{x-\mu}{\sigma})^2)
    \label{eq:NormalDistribution}
\end{equation}
\begin{equation}
    q(x_t | x_{t-1}) = \mathcal{N}(x_t; \sqrt{1-\beta_t}x_{t-1}, \beta_t\textbf{I})
    \label{eq:ForwardDiffusion_q}
\end{equation}
\begin{equation}
    F(x_0, \beta) = \prod_{t=1}^T q(x_t | x_{t-1})
    \label{eq:ForwardDiffusion}
\end{equation}
where 
$x_0$ denotes to the original image, $\beta$ refers to a variance schedule with timesteps $t$, $T$ represents the total number of timesteps, and $x_t$ refers to the original image noised to the degree of timestep $t$. 

After forward diffusion, a generative model is used to learn the reverse diffusion process. 
The reverse diffusion process is described as:
\begin{equation}
    p_\theta(x_{t-1} | x_t) = \mathcal{N}(x_{t-1}; \mu_\theta(x_t, t), \beta_t\textbf{I})
    \label{eq:BackwardDiffusion_p}
\end{equation}
\begin{equation}
    p_\theta(x_{0:T}) = \mathcal{N}(x_T; \textbf{0}, \textbf{I}) \prod_{t=1}^T p_\theta(x_{t-1} | x_t)
    \label{eq:BackwardDiffusion}
\end{equation}
where
$x_T$ denotes a fully noised image with $\mu_\theta$ being the learnable parameters of the model.
The loss function for a given timestep $t$ is a simple $L_1$ loss between the generated image and the original one as: 
\begin{equation}
    L_1 = |x_0 - \Tilde{x_0}|
    \label{eq:diffusion_loss}
\end{equation}

\begin{figure}
  \centering
  \includegraphics[width=0.475\textwidth]{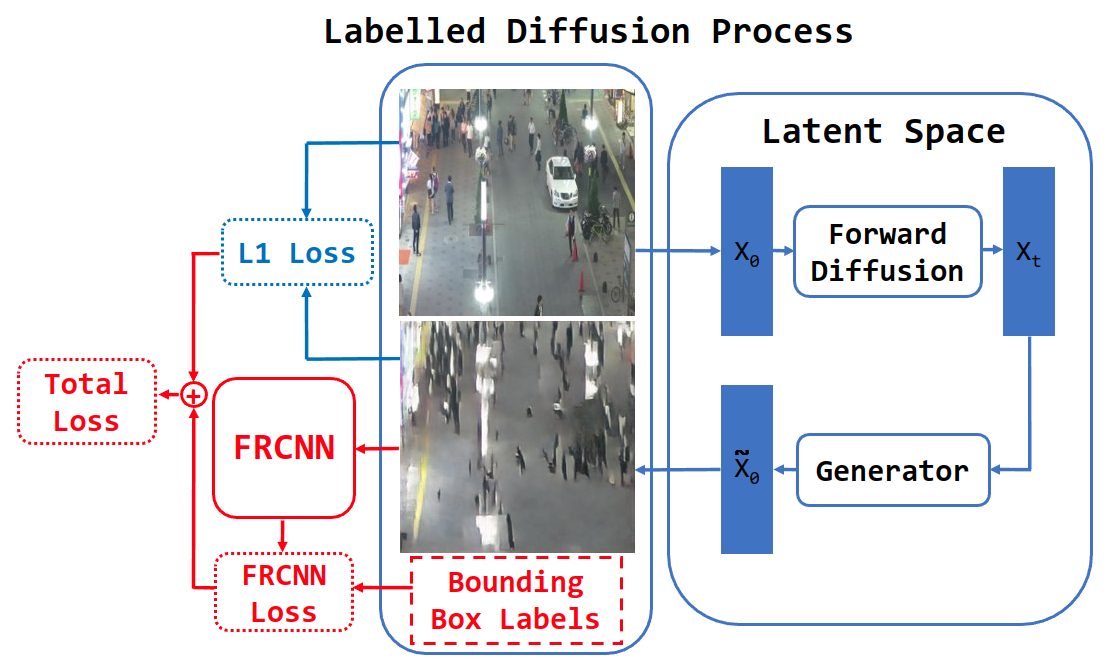}
  \caption{The basic diffusion architecture (blue), and the
  modified diffusion architecture for generating images and corresponding labels (red+blue)}
  \label{fig:diffusion_arch}
\end{figure}

The overall process is outlined in Figure \ref{fig:diffusion_arch}. This is a simplified explanation of the diffusion training process. The work by Cao et al.~\cite{ddpm_review} explains this process in more depth. 

\section {Method}



The proposed method 
uses a diffusion model to generate a dataset which 
improves training of pedestrian detection models.
During training, the input to the pedestrian detection model is a series of images 
with ground truth pedestrian bounding boxes. 
At inference, 
the pedestrian detection model
outputs a set of bounding boxes which are predictions of pedestrian locations.
Our dataset generation process 
during training
accepts as input
a dataset of images with corresponding bounding boxes, 
and, on inference, outputs 
a set of generated synthetic labelled images
with a distribution close to that of the input dataset.
This synthetic dataset is then used to
augment a target dataset during training
of the pedestrian model,
to boost performance.

\subsection{Diffusion Dataset Generation}

 To augment and improve simulated data, a new data generation method is proposed. In this method, diffusion models are used to learn and generate data in the real domain (i.e. CityPersons, MOT17) while simultaneously learning to label these images. The proposed pipeline comprises two stages. First, a modified diffusion architecture learns to produce images in its trained domain while also learning to produce bounding boxes for the pedestrians in generated images. Then, this trained model is used to produce an arbitrarily sized dataset through sampling and the generated annotations are converted to the appropriate format for pedestrian detection training. This section outlines the architectures and processes for these components while the general pipeline can be viewed in Figure \ref{fig:general_pipeline}.

\subsubsection{Diffusion Architecture}
While standard diffusion models are able to produce extra images for a real dataset, they do not natively support the labelling of these images. To provide labels for the generated images a modified diffusion architecture is proposed. This architecture appends an Faster R-CNN (FRCNN) module to the end of the diffusion process. This module uses the bounding boxes provided by the ground truth of the real dataset coupled with the result of backwards diffusion to learn how to label generated images while the diffusion model learns to produce them. The loss function for FRCNN is added to the loss of diffusion to provide a total, coupled loss function allowing simultaneous training of these models. This architecture can be seen in Figure \ref{fig:diffusion_arch}. Once this model is trained, it is able to generate images in the domain it was trained on while providing a series of bounding boxes for each image which represent the pedestrians in each image.


\subsection{Pedestrian Detection}
\label{sec:ped_det_method}
As this paper focuses on improving the results provided by MOTSynth, we use their codebase to perform our experiments. MOTSynth experiments with training on their simulated data using four different object detection methods: YOLOv3, CenterNet, Faster R-CNN, and Mask R-CNN. Due to our use of Faster R-CNN in diffusion generation, we experiment exclusively with Faster R-CNN as our object detection method. 

MOTSynth describes four different subsets of data which they use for training and testing. These four subsets include 72, 144, 288, and 576 video clips of data and are labelled MOTSynth-1, MOTSynth-2, MOTSynth-3, and MOTSynth-4 respectively. Unfortunately, due to hardware restrictions, we were unable to train on this quantity of data. Instead, we utilize three different subsets which contain 1, 4, and 61 videos clips denoted MOTSynth-A, MOTSynth-B, and MOTSynth-C respectively. 

\subsection{Dataset Augmentation Method}
After training separate diffusion models for MOT17 and CityPersons, two generated datasets for each real dataset are created containing 1,000 and 10,000 images. These are denoted as MOT17 1k, MOT17 10k, CityPersons 1k, and CityPersons 10k. It should be noted that our diffusion generation method only produces images at a resolution of 256x256 while MOTSynth, MOT17, and CityPersons have resolutions of 1920x1080, 1920x1080 and 640x480, and 2048x1024 respectively. To account for this, we run two sets of experiments. First, to augment MOTSynth A through C, we combine each generated dataset, at a resolution of 256x256, with each synthetic dataset at their original resolutions which will be evaluated on MOT17 at full resolution. Then, we repeat this combination with the 256x256 representation of MOTSynth A through C which will be evaluated on MOT17 at a resolution of 256x256. The result of these combinations is 24 augmented datasets that we can use to compare against the baseline achieved by training MOTSynth A through C on their own. In this way we can determine if our method results in an improvement of AP.

\section{Experiments and Results}
\subsection{Real-World Pedestrian Detection Datasets}
\begin{table} 
\centering
\caption{The size and annotations of CityPersons and MOT17}
\begin{tabular}{c|c|c}
    \hline
    \textbf{Dataset} & \textbf{CityPersons~\cite{citypersons}} & \textbf{MOT17~\cite{MOT17}} \\
    \hline
    \# Frames & 3k & 11k \\
    \# Instances & 11k & 292k \\
    3D & - & - \\
    Pose & - & - \\
    Segmentation & \cmark & - \\
    Depth & \cmark & - \\
    \hline
\end{tabular}
\label{table:ped_dataset-stats}
\end{table}
There exist multiple real-world pedestrian detection datasets collected in different environments. Specifically, the two focused on in this paper are CityPersons~\cite{citypersons} and MOT17~\cite{MOT17}. CityPersons is a subset of the urban scene dataset known as CityScapes~\cite{cityscapes}. It contains street scenes, collected from a dashcam view, which display pedestrians on the streets of different European cities~\cite{citypersons}. 

MOT17 is a dataset from the MOTChallenge, an annual challenge focused on the tasks of object detection, pedestrian detection, 3D reconstruction, optical flow, single-object short-term tracking, and stereo estimation~\cite{MOT17}. This dataset consists of seven indoor and outdoor public scenes with pedestrians as the focus~\cite{MOT17}. All images are collected from a stationary video camera. The statistics and sizes of these datasets can be viewed in Table \ref{table:ped_dataset-stats}.

In their experiments,~\cite{motsynth} uses Average Precision (AP) as their ultimate evaluation metric. Following this work, we also use AP to compare models with a minimum IoU of 0.5.

To utilize the proposed diffusion architecture, these real-world datasets are looked to for diffusion training. First, the MOT17 dataset is used. This is the dataset that~\cite{motsynth}, as well as our paper, uses as the real-world evaluation benchmark. In~\cite{motsynth}, they train a pedestrian detection model on multiple subsets of MOTSynth, and then test on MOT17 to see how well the model generalizes the synthetic data for real-world application. To ensure that diffusion is able to produce images that will improve training we use MOT17 to train our diffusion model and produce extra images for MOTSynth. Next, we look to a separate pedestrian detection dataset to see if other real-world data is able to improve MOTSynth results through augmentation. The dataset chosen is CityPersons which has one significant difference when compared to MOTSynth and MOT17. CityPersons, while still labelling pedestrians, collected data from a dashcam view of a car rather than a stationary overhead or sidewalk camera. We use CityPersons to train a diffusion model for augmentation to see if our method will still improve results when trained on a dataset that is distinct from the real-world evaluation benchmark.

\begin{figure*}
  \centering
  \includegraphics[width=\textwidth]{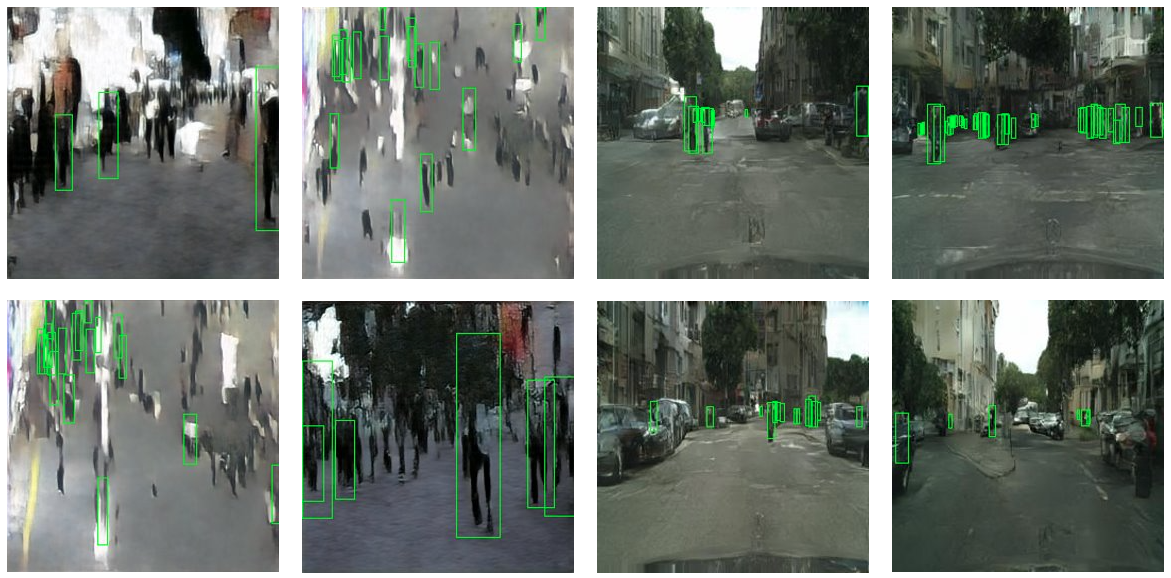}
  \caption{MOT17 diffusion generated images (left). CityPersons diffusion generated images with modified pipeline (right).}
  \label{fig:MOT17_CP_diffusion_comparison}
\end{figure*}

Through experimentation, it was found that the diffusion model converged much easier on semantic segmentation maps rather than the original RGB images. As CityPersons has readily available semantic segmentation maps as part of the dataset, the data generation pipeline changed. We used diffusion to learn how to produce CityPersons semantic segmentation maps, while is also learned to label these images. Then, we generated a dataset from the trained model and use a pix2pix GAN-based model to transfer these generated semantic images into the real CityPersons domain. The addition of image synthesis in the pipeline can be seen in Figure \ref{fig:overall_pipeline}. Comparing the images generated by this method compared to the ones generated by directly training on MOT17, we see a large increase in image quality which can be viewed in Figure \ref{fig:MOT17_CP_diffusion_comparison}. We did not replicate this method for MOT17 due to its lack of semantic segmentation maps.

\subsection{MOTSynth Baselines} 

\begin{table} 
\centering
\caption{Average precision results from training on MOTSynth A through C trained on the given dataset and tested on MOT17 at the corresponding resolution as well as the MOTSynth reported results for datasets MOTSynth 1 through 4. Included is the size of each dataset in number of video clips}
\begin{tabular}{lcccc}
    \textbf{} & \multicolumn{2}{c}{\textbf{Resolution}} & & \\
    \textbf{Dataset:} & \textbf{1920x1080} & \textbf{256x256} & \textbf{\# Clips} & \textbf{\# Frames} \\
    \hline
    MOTSynth-A & 49.6\% & 23.0\% & 1 & 2k \\
    MOTSynth-B & 64.0\% & 38.9\% & 4 & 7k \\
    MOTSynth-C & \textbf{71.2\%} & \textbf{43.8\%} & 61 & 110k \\
    \hline
    MOTSynth-1 & 76.8\% & - & 72 & 130k \\
    MOTSynth-2 & 77.5\% & - & 144 & 259k \\
    MOTSynth-3 & 78.3\% & - & 288 & 518k \\
    MOTSynth-4 & \textbf{79.0\%} & - & 576 & 1037k \\
    \hline
\end{tabular}
\label{table:motsynth-baselines}
\end{table}

Before determining if our diffusion generated data improved the results from~\cite{motsynth}, we first determined if we could repeat their baseline results. Additionally, our diffusion method produced images at a resolution of 256x256 while MOTSynth data has a resolution of 1920x1080. To address this, we collected the baselines for both MOTSynth and MOT17, scaled to 256x256. Our obtained baselines for the datasets described in Sec.~\ref{sec:ped_det_method}, along with their smaller representations and the reported MOTSynth results, can be seen in Table \ref{table:motsynth-baselines}.

There are two key insights from our obtained baselines. 
First, the AP results that we obtained on our partitions of MOTSynth were
lower than those reported in the original paper. This was expected as our training dataset partitions were smaller than those used previously. For example, MOTSynth-C with 61 clips performed 5.6\% lower than MOTSynth-1 with 72 clips, which indicates similar performance.
Next, overall results dropped 
when training and testing on the smaller 256x256 image resolution. 
Due to the discrepancy between our MOTSynth results and the reported results, along with the drop in AP when training and testing on the smaller resolution, 
we 
considered
the relative improvement we were able to achieve compared against our collected baselines
as our measure of success, rather than the absolute AP. 


\subsection{Diffusion Dataset Augmentation}

We experimented with augmenting our simulated data with MOT17-like and CityPersons-like diffusion generated data. In this way, we determined if our method provided an increase in performance and how large that increase was when using generated data which represented two different datasets.

\begin{table}
\centering
\caption{Average precision results from training on MOTSynth A through C, augmented with MOT17 1k and MOT17 10k, trained on the given dataset and tested on MOT17 at the corresponding resolution}
\begin{tabular}{lccc}
     &  & \multicolumn{2}{c}{\textbf{Resolution}}\\
    
    \textbf{Base Dataset} & \textbf{Augmentation} & \textbf{1920x1080} & \textbf{256x256}\\
    \hline
     & None & 49.6\% & 23.0\% \\
    MOTSynth-A & MOT17 1k & \textbf{53.3\%} & \underline{34.4\%} \\
     & MOT17 10k & \underline{51.7\%} & \textbf{50.3\%} \\
    \hline
     & None & 64.0\% & 38.9\% \\
    MOTSynth-B & MOT17 1k & \textbf{66.0\%} & \underline{47.9\%} \\
     & MOT17 10k & \underline{65.3\%} & \textbf{57.4\%} \\
    \hline
     & None & 71.2\% & 43.8\% \\
    MOTSynth-C & MOT17 1k & \underline{72.1\%} & \underline{52.9\%} \\
     & MOT17 10k & \textbf{72.5\%} & \textbf{57.9\%} \\
    \hline
\end{tabular}
\label{table:motsynth-mot17aug}
\end{table}

We initially aimed to gauge the feasibility of our method, by 
generating images based on the
MOT17 test partition.
In this way, the generated images
would be as close as possible to the test images,
as they were trained from the same population.


As all pedestrian detection models trained with synthetic data were tested on MOT17, we expected a marked improvement in the results by augmenting our training data with generated MOT17-like images. 

To experiment with augmentation, we generated two datasets with 1k and 10k images denoted MOT17 1k and MOT17 10k respectively. The results from this experiment can be seen in Table \ref{table:motsynth-mot17aug}.

When comparing the
results of the full resolution MOTSynth subsets,  
MOTSynth-C provided the best overall AP of 72.5\% when augmented with MOT17 10k which is a 1.3\% improvement over the MOTSynth-C baseline. 
As the amount of MOTSynth data decreases, we observed a larger improvement over the baselines of 3.7\% and 2.0\% for MOTSynth-A and MOTSynth-B respectively. Interestingly, these improvements came from MOT17 1k rather than the larger 10k generated dataset. We postulate that this is due to the smaller dataset sizes of MOTSynth A and B and training on the larger 10k generated dataset caused overfitting on the smaller image resolution generated data.

Moving on to the smaller resolution MOTSynth subsets, we observed a much larger improvement when augmenting with diffusion generated data. This was expected as the generated data matches MOTSynth in image resolution so the model will not be confused by the difference in image sizes. Again, the best result came from MOTSynth-C augmented with MOT17 10k which achieved an AP of 57.9\%, a 14.1\% improvement over the baseline. The trend of smaller MOTSynth subsets corresponding to a larger improvement with augmentation falters slightly here. MOTSynth-A, augmented with MOT17 10k, produces our largest improvement of 27.3\% while MOTSynth-B, augmented with MOT17 10k, produces an improvement of 18.5\%. It appears that using more generated data leads to better results when the resolutions are matched. This is expected as the generated data, used here for augmentation, is made to match the evaluation dataset.

\begin{table}
\centering
\caption{Average precision results from training on MOTSynth A through C, augmented with CityPersons 1k and CityPersons 10k, trained on the given dataset and tested on MOT17 at the corresponding resolution}
\begin{tabular}{lccc}
    \textbf{} &  & \multicolumn{2}{c}{\textbf{Resolution}}\\
    
    \textbf{Base Dataset} & \textbf{Augmentation} & \textbf{1920x1080} & \textbf{256x256}\\
    \hline
     & None & 49.6\% & 23.0\% \\
    MOTSynth-A & CityPersons 1k & \textbf{52.5\%} & \underline{34.5\%} \\
     & CityPersons 10k & \underline{50.3\%} & \textbf{36.8\%} \\
    \hline
     & None & 64.0\% & 38.9\% \\
    MOTSynth-B & CityPersons 1k & \underline{64.6\%} & \textbf{45.0\%} \\
     & CityPersons 10k & \textbf{66.0\%} & \underline{43.7\%} \\
    \hline
     & None & 71.2\% & \underline{43.8\%} \\
    MOTSynth-C & CityPersons 1k & \underline{72.6\%} & 41.6\% \\
     & CityPersons 10k & \textbf{73.0\%} & \textbf{48.1\%} \\
    \hline
\end{tabular}
\label{table:motsynth-CPaug}
\end{table}

Next, to determine if we are able to improve results with a separate real-world dataset, we train a diffusion model on CityPersons to produce data for augmentation. Should there be an improvement in AP we expect it to be less than augmentation with MOT17-like data as CityPersons is different than the MOT17 evaluation benchmark. Again, we generate two datasets with 1k and 10k images denoted CityPersons 1k and CityPersons 10k respectively. The results from this experiment can be seen in Table \ref{table:motsynth-CPaug}.


Beginning with the full resolution MOTSynth subsets, we achieved the best AP of 73.0\% on MOTSynth-C augmented with CityPersons 10k which is an improvement of 1.8\% on the baseline. Unexpectedly, this AP is better than MOTSynth-C augmented with MOT17 10k. 
There are two reasons why this might be the case, one being due to the lower image quality of MOT17 10k when compared to CityPersons 10k. 
The second reason might be the slight difference in the pipelines used to generate the images,
which derive from the datasets themselves,
as described above and shown in Fig.~\ref{fig:MOT17_CP_diffusion_comparison}.

In contrast with the previous results,
here the smaller subsets of MOTSynth do not show a larger improvement with augmentation.
When augmenting with CityPersons 1k the improvements are 2.9\%, 0.6\%, and 1.4\% for MOTSynth A through C respectively. Additionally, when augmenting with CityPersons 10k, the improvements are 0.7\%, 2.0\%, and 1.8\% for MOTSynth A through C respectively. 


Looking at the smaller resolution MOTSynth subsets, we achieved the best AP of 48.1\% on MOTSynth-C augmented with CityPersons 10k; a 4.3\% improvement over the baseline. Interestingly, we find the first case of diffusion generated data decreasing performance when compared to the baseline. MOTSynth-C augmented with CityPersons 1k showed a 2.2\% reduction in AP which may be due to the class imbalance resulting from the relatively small augmented dataset. Again, the general trend with smaller resolution datasets shows the biggest improvement when more generated data is used. While this trend breaks when augmenting MOTSynth-B, we believe this is due to the domain discrepancies between CityPersons and MOT17.

In almost every experiment run here, we have shown that augmenting simulated data with diffusion generated data improves the AP when compared to models trained exclusively on simulation. These results seem to hold up regardless of dataset size, simulated vs. generated split, and image resolution. In a more fair comparison, where the image resolution of both simulated, generated, and real-world evaluation data is equal, we showed our largest jump in AP (from 23.0\% to 50.3\%). This suggests our method works effectively and would show a much larger improvement in AP on larger resolutions should the size of our generated data increase.


\section{Conclusion}
This work presents a novel diffusion pipeline to generate real-world data for simulated dataset augmentation. We show that we are consistently able to outperform pedestrian detection models that are trained on purely simulated data using this technique with our best improvement being 27.3\%. 

We view two ways in which our technique could be improved. 
First, our method generates images at a resolution of 256x256 while the synthetic dataset we are augmenting has a resolution of 1920x1080. Improving our diffusion model to generate images at the full resolution should improve the pedestrian detection model performance significantly as we showed in our experiments on the 256x256 versions of MOTSynth and MOT17. 
Second, there were some missed labels in the generated data. Experimenting with different,
more powerful object detectors leveraged by diffusion models, or simply training for longer, could improve the annotations generated leading to an increase in performance.
Our generated datasets can be viewed here: \url{https://github.com/And1210/Diffusion-Generated-Datasets}.


\section*{Acknowledgment}
Thanks to Geotab Inc., the City of Kingston, and NSERC for their support of this work.


\bibliographystyle{plain}
\bibliography{references}




\end{document}